\documentclass[letterpaper, 10 pt, conference]{ieeeconf}  

\IEEEoverridecommandlockouts                              

\usepackage{amsmath, amssymb,bbm,bm,mathtools}
\usepackage[ruled,vlined]{algorithm2e}
\usepackage{comment}
\usepackage{url}

\usepackage{graphics} 
\usepackage{epsfig} 
\usepackage{amsmath} 
\usepackage{amssymb}  
\usepackage{textcomp}

\DeclareMathOperator*{\diag}{diag} 
\DeclareMathOperator*{\blkdiag}{blkdiag} 
\renewcommand{\Re}{\mathbb{R}}
\newcommand{\X}{\mathbf{x}}
\newcommand{\U}{\mathbf{u}}
\newcommand{\V}{\mathbf{v}}
\newcommand{\Y}{\mathbf{y}}
\newcommand{\D}{\mathbf{d}}
\newcommand{\W}{\mathbf{w}}
\newcommand{\e}{\bm{\epsilon}}

\renewcommand{\baselinestretch}{0.98}

\title{\LARGE \bf
Trajectory Distribution Control for Model Predictive Path Integral Control using Covariance Steering
}

\author{Ji Yin, Zhiyuan Zhang, Evangelos Theodorou, Panagiotis Tsiotras
\thanks{The authors are with the D. Guggenheim School of Aerospace Engineering, 
Georgia Institute of Technology, GA. E-mail: {\tt\small \{jyin81, nickzhang, evangelos.theodorou, tsiotras\}@gatech.edu}}%
}

\begin{document}

\maketitle
\thispagestyle{empty}
\pagestyle{empty}

\begin{abstract}
    This paper presents a novel control approach for autonomous 
    systems operating under uncertainty. 
    We combine Model Predictive Path Integral (MPPI) control with Covariance Steering (CS) theory to obtain a robust controller for general nonlinear systems. 
    The proposed Covariance-Controlled Model Predictive Path Integral (CC-MPPI) controller addresses the performance degradation observed in some MPPI implementations owing to unexpected disturbances and uncertainties. 
    Namely, in cases where the environment changes too fast or the simulated dynamics during the MPPI rollouts do not capture the noise and uncertainty in the actual dynamics, the baseline MPPI implementation may lead to divergence. 
    The proposed CC-MPPI controller avoids divergence by controlling the dispersion of the rollout trajectories at the end of the prediction horizon. Furthermore, the CC-MPPI has adjustable trajectory sampling distributions that can be changed according to the environment to achieve efficient sampling.
    Numerical examples using a ground vehicle navigating in challenging environments demonstrate the proposed approach.
    \end{abstract}

\section{Introduction}

    As autonomous vehicles and other robots become increasingly popular, one of the major concerns is whether humans can trust the robots' ability to complete their assigned tasks safely. 
    For autonomous vehicles, for example, neglecting or misinterpreting the disturbances during driving can lead to serious consequences within milliseconds. 
    The demand for safety has led many researchers develop robust control and planning algorithms for autonomous robotic systems. For example, sampling-based planning algorithms that consider uncertainty and collision probability in the vertex selection or evaluation processes have been proposed in \cite{SafeRRT, ParicleRRT, CollisionProbabilityPRM, StochasticUncertainRRT}, and optimization-based planning algorithms that consider systematic disturbances and chance constraints explicitly by solving optimization problems have been developed in \cite{BakolasCS, MaximCS, CSvehicle, CS}. 
    
    In this paper, we propose an MPC-based robust trajectory planning approach that deals with environmental and plant uncertainty, while providing guarantees on the dispersion of the \textit{closed-loop} system future trajectories.
     Model Predictive Control (MPC) is an algorithmic, optimization-based control design \cite{AlgorithmicControlTheory} that has gained popularity for autonomous vehicle control over the past years \cite{Rosolia_Borrelli,Racing_miniature_cars}. 
     The traditional deterministic MPC approaches are model-based and generate trajectories assuming there are no uncertainties in the dynamics.
     As a result, MPC controllers are typically not robust to model parameter variations. 
     To improve the performance of MPC controllers by taking system uncertainty into account, Robust MPC (RMPC) controllers have been proposed to handle deterministic uncertainties residing in a given compact set. 
     RMPC generates control commands by considering worst-case scenarios, thus the resulting trajectories can be conservative. 
     Reference~\cite{RMPC} provides an extensive review summarizing all types of RMPC controllers. 
     To achieve more aggressive planning, Stochastic MPC (SMPC) utilizes the probabilistic nature of the system uncertainty to account for the most likely disturbances, instead of considering only the worst-case disturbance, as with the RMPC \cite{SMPC1,SMPC2}. There are two classes of SMPC approaches in the literature. The first one is based on the analytical solutions of some optimization problem, such as \cite{PTMPC,PCMPC,CSSMPC}, while the second approach relies on randomization to solve optimization problems, such as \cite{SBMPC1,SBMPC2,firstMPPI}. The proposed CC-MPPI controller is somewhere in-between these two, as it analytically computes a controlled dynamics by considering the model uncertainty and then generates the optimal control using randomized roll-outs of the controlled dynamics. This is discussed in greater detail in Section \ref{Covariance Controlled MPPI}.
        
    Most of current MPC implementations assume linear system dynamics and formulate the resulting MPC task as a quadratic optimization problem, which helps MPC meet the strict real-time requirements required for safe control. However, these approaches depend on simplified linear models that may not capture accurately the dynamics of the real system. 
    Model Predictive Path Integral (MPPI) control \cite{Information_theoretic_MPC} is a type of
    nonlinear MPC algorithm that solves repeatedly finite-horizon optimal control tasks while utilizing nonlinear dynamics and general cost functions. 
    Specifically, MPPI is a simulation-based algorithm that samples thousands of trajectories around some mean control sequence in real-time, by taking advantage of the parallel computing capabilities of modern Graphic Processing Units (GPUs). 
    It then produces an optimal trajectory along with its corresponding control sequence by calculating the weighted average of the cost of the ensuing sampled trajectories, where the weights are determined by the cost of each trajectory rollout. 
    %
    One of the advantages of the MPPI approach over more traditional MPC controllers is that it does not restrict the form of the cost function of the optimization problem \cite{MPPI_Theory_and_Applications_to_Autonomous_Driving}, which can be non-quadratic and even discontinuous.
        
\begin{figure}[t]
    \centerline{\includegraphics[scale = 0.25]{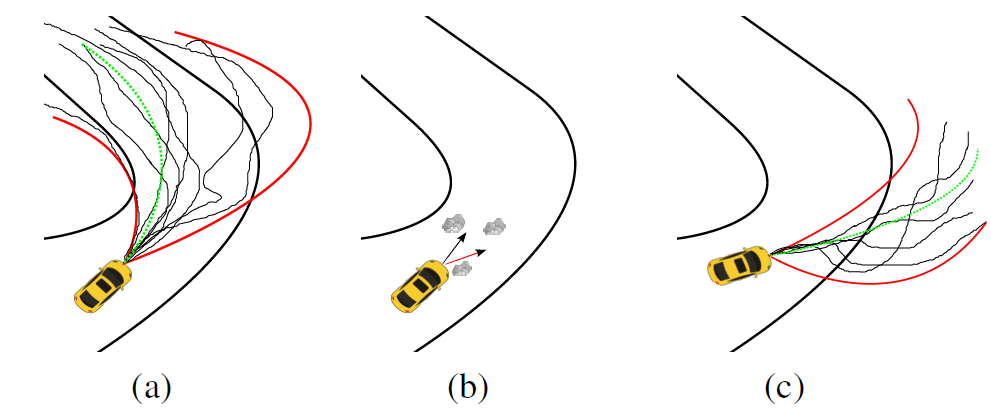}}
    \caption{MPPI Divergence; from~\cite{ETRSS2018}.} 
    \label{fig1}
    \end{figure}

    Despite its appealing characteristics, the MPPI algorithm may encounter problems when implemented in practice. 
    In particular, when the mean control sequence lies inside an infeasible region,  all the resulting MPPI sampled trajectories are concentrated within the same region, as illustrated in Figure \ref{fig1}, and 
    this may lead to a situation where the trajectories violate the constraints.
    Two cases this may happen are: first, when the MPPI algorithm diverges because the environment changes too fast; and, second, when the algorithm fails because the predicted dynamics do not capture the noise and uncertainty of the actual dynamics. 
    The reason the MPPI algorithm may perform poorly under the previous two cases is because it fails to take into account the disturbances (either from the dynamics or from the environment) so that all sampled trajectories end up violating the constraints. 
    Figure~\ref{fig1} shows the
    influence of the noise on MPPI sampled trajectories. 
    In this figure, the gray curves are the MPPI sampled trajectories,  the red curves show the boundaries of the trajectory sampling distribution, and the green curve represents the simulated trajectory of the robot following the optimal control sequence given the current distribution.
    In Figure~\ref{fig1}(a)
    the autonomous vehicle has sampling distribution mostly inside the track initially.
    In Figure~\ref{fig1}(b), the vehicle ends up in an unexpected pose due to unforeseen disturbances after it executes the control command.
    This further leads to the situation depicted in Figure~\ref{fig1}(c), 
    where the algorithm diverges because all of the sampled trajectories violate the constraints.
         
    To mitigate the previous  shortcomings of the MPPI algorithm, prior works apply 
    a  controller to track the output of the MPPI controller in order to keep the actual trajectory as close as possible to the predicted nominal trajectory. 
    These approaches separate the planning and control tasks so that MPPI acts similarly to a path planner. 
    For example, in \cite{ETRSS2018} an iterative Linear Quadratic Gaussian (iLQG) controller  was used to track the planned trajectory provided by MPPI. 
    In~\cite{L1AdaptiveMPPI} the authors propose a method that utilizes a tracking controller with $L_1$ augmentation to compensate for the mismatch between the nominal dynamics and the true dynamics. 
    However, these methods do not improve the performance of the MPPI algorithm if there are significant changes in the environment within a short interval of time. 
    The proposed CC-MPPI algorithm tries to address some of these shortcomings by improving the performance of the MPPI algorithm under the scenarios mentioned above. 
    This is achieved by introducing adjustable trajectory sampling distributions, and by directly controlling the evolution of these trajectory distributions to avoid an uncontrolled dispersion at the end of the control horizon.

\section{Problem Formulation}\label{Problem Formulation}
    
   The goal of the proposed Covariance-Controlled MPPI (CC-MPPI) controller is to make the distributions of the sampled trajectories more flexible than the ones generated by MPPI, such that the CC-MPPI algorithm samples more efficiently and with a smaller probability to be trapped in local minima when the optimal trajectory from the previous time step lies inside some high-cost region, as illustrated in Figure~\ref{fig1}(c). 
    To this end, we introduce a desired terminal state covariance $\Sigma_f$ for the states of the dynamics (\ref{NonlinearSystem}) at the final time step $N$ as a hyperparameter for the CC-MPPI controller. 
    The key idea is that the distribution of the sampled trajectories can be adjusted by a suitable choice of $\Sigma_f$ together with the control disturbance variance $\Sigma_{\epsilon}$. 
    %
%
    Denoting by $\V = \left[v_0, \ldots, v_{N-1}\right]^\intercal \in \mathbb{R}^{n_u N}$,
    the CC-MPPI controller solves the following optimization problem,
    \begin{subequations}\label{Problemformulation}
    \begin{align}
        \begin{split}
        \min_\V J(\V) &= \\
        &\mathbb{E}\left[ \phi(x_N) + \sum^{N-1}_{k=0} \left(q(x_k)  + \frac{1}{2}v_k R v_k \right) \right],\label{CC-MPPI objective}
        \end{split}\\
    \text{subject to,} \nonumber\\
    &x_{k+1} = F(x_k, v_k + \epsilon_k),\label{NonlinearSystem}\\
    &x(0) = x_0, \quad \epsilon_k \sim \mathcal{N}(0, \Sigma_{\epsilon}),\label{eq1c}\\
    &\mathbb{E}\left[x_N{x_N}^\intercal\right] - \mathbb{E}\left[x_N\right]\mathbb{E}\left[x_N\right]^\intercal \preceq \Sigma_f, \label{terminalconstraint}
    \end{align}
    \end{subequations}
    at each iteration, where the state terminal cost $\phi(x_N)$ and the state portion of the running cost $q(x_k)$ can be arbitrary functions. The objective function (\ref{CC-MPPI objective}) minimizes the expectation of the state and control costs with $x_k$ being a random vector subject to the dynamics (\ref{NonlinearSystem}). 

\section{MPPI Algorithm Review}\label{MPPI Algorithm Review}

    The MPPI controller, as described in \cite{MPPI_Theory_and_Applications_to_Autonomous_Driving},
    minimizes (\ref{CC-MPPI objective}) subject to (\ref{NonlinearSystem}) and (\ref{eq1c}).
    As in Problem (\ref{Problemformulation}), the terminal cost $\phi(x_N)$ and the state portion of the running cost $q(x_k)$ of the MPPI can be arbitrary functions.
    
   As in an MPC setting, the MPPI algorithm samples $M$ trajectories during each optimization iteration. 
   Let $\V = \left[v_0, \ldots, v_{N-1}\right]^\intercal \in \mathbb{R}^{n_u N}$ be the mean control sequence, $\U^{(m)} = \left[u^{(m)}_0, \ldots, u^{(m)}_{N-1}\right]^\intercal \in \mathbb{R}^{n_u N}$ be the actual control sequence, and let $\e^{(m)} = \left[\epsilon_0^{(m)}, \ldots,\epsilon_{N-1}^{(m)}\right]^\intercal \in \mathbb{R}^{n_u N}$, $\epsilon^{(m)}_k \sim \mathcal{N}(0,\Sigma_{\epsilon})$, be the control disturbance sequence corresponding to the $m^{\text{th}}$ sampled trajectory $\left[ x^{(m)}_0, \hdots, x^{(m)}_{N-1} \right]$ at the current iteration, such that $\U^{(m)} = \V + \e^{(m)}$, where $m = 1,\ldots,M$.  
   The cost for the $m^{\text{th}}$ sampled trajectory is given by \cite{firstMPPI}
    \begin{equation}\label{eq2}
        S_m = \phi(x^{(m)}_N) + \sum^{N-1}_{k=0}\Tilde{q}(x^{(m)}_k,v_k,\epsilon^{(m)}_k),
    \end{equation}
     where $\Tilde{q}(x^{(m)}_k,v_k,\epsilon^{(m)}_k)$ is the cost for the $m^{\textrm{th}}$ sampled trajectory at the $k^{\textrm{th}}$ step, given by
    \begin{equation}\label{ModifiedCost}
        \begin{split}
            &\Tilde{q}(x^{(m)}_k, v^{(m)}_k, \epsilon^{(m)}_k) = q(x^{(m)}_k) +\\
            & \frac{1-\nu^{-1}}{2}{\epsilon^{(m)}_k}^\intercal R \epsilon^{(m)}_k + {v^{(m)}_k}^\intercal R \epsilon^{(m)}_k + \frac{1}{2} {v^{(m)}_k}^\intercal R {v^{(m)}_k},        
        \end{split}
    \end{equation}
    where $\nu$ is the ratio between the covariance of the injected disturbance $\e^{(m)}$ and the covariance of the disturbance of the original dynamics. 
    The term $\frac{1}{2}{v^{(m)}_k}^\intercal R {v^{(m)}_k}$ in (\ref{ModifiedCost}) is the cost for the disturbance-free portion of the control input, and both $\frac{1-\nu^{-1}}{2}{\epsilon^{(m)}_k}^\intercal R \epsilon^{(m)}_k$ and ${v^{(m)}_k}^\intercal R \epsilon^{(m)}_k$ in (\ref{ModifiedCost}) penalize large control disturbances and smooth out the resulting control signal. The weights $\omega_m$ $(m = 1,\ldots,M)$ of the $m^{\text{th}}$ sampled trajectory are chosen as \cite{MPPIracing}
    \begin{equation}\label{eq3}
        \omega_m = \text{exp}\left(-\frac{1}{\lambda}\left(S_m - \beta \right)\right),
    \end{equation}
    where,
    $
        \beta = \min_{m=1,\ldots,M} S_m,
    $
    and where $\lambda$ determines how selective is the weighted average of the sampled trajectories. Note that the constant $\beta$ does not influence the solution, and it is introduced to prevent numerical instability of the algorithm. 
    The MPPI algorithm generates the optimal control sequence and the mean sequence of the next iteration $\V$ using the following equations,
    \begin{equation}\label{eq5}
    \V = \sum^{M}_{m=1}\omega_m \U^{(m)}/ \sum^{M}_{m = 1}\omega_m.
    \end{equation}
    In Section \ref{Covariance Controlled MPPI} we discuss how the CC-MPPI controller satisfies the terminal constraint (\ref{terminalconstraint}), and then present the complete CC-MPPI algorithm. 
    
\section{Covariance-Controlled MPPI} \label{Covariance Controlled MPPI}

    In this section, we introduce the proposed CC-MPPI controller. 
    Section~\ref{Linearized Model} discusses the linearization of the dynamics (\ref{NonlinearSystem}), and 
    Section~\ref{Covariance controlled trajectory sampling} uses this linearization to achieve the terminal constraint (\ref{terminalconstraint}). 
    Section~\ref{The CC-MPPI Algorithm} presents the proposed CC-MPPI algorithm that solves the optimization Problem (\ref{Problemformulation}).
    
    \subsection{Linearized Model}\label{Linearized Model}
    
    We start by linearizing system (\ref{NonlinearSystem}) along some reference trajectory using the approach outlined in \cite{LTVlinearization}. 
    The reference trajectory of the first optimization iteration can be an arbitrary trajectory; 
    starting with the second iteration, the reference trajectory of the current iteration is the trajectory generated by the optimal control sequence from the previous iteration. 
    To this end, let $\W^r = \left[w^r_0,\ldots,w^r_{N-1}\right]^\intercal$ be the reference control sequence at the current iteration, and let $\X^r = \left[x^r_0,\ldots,x^r_N\right]^\intercal$ be the corresponding reference state sequence, such that
    \begin{equation}\label{eq7}
    x^r_{k+1} = F(x^r_k,w^r_k),\qquad k = 0,\ldots,N-1.
    \end{equation}
    The dynamical system in (\ref{eq7}) can then be approximated in the vicinity of $(\X^r, \W^r)$ with a discrete-time, linear time-varying (LTV) system as follows
    \begin{equation}\label{eq8}
    x_{k+1} = A_kx_k + B_ku_k + d_k,
    \end{equation}
    where $x_k$ and $u_k$ are the state and control input respectively to the LTV system at step $k$, and
    \begin{equation}\label{eq9}
    A_k = \frac{\partial F}{\partial x_k}\Bigr|_{x_k = x^r_k,u_k=w^r_k}, \quad B_k = \frac{\partial F}{\partial u_k}\Bigr|_{x_k = x^r_k,u_k=w^r_k},
    \end{equation}
    \begin{equation}\label{eq10}
    d_k = F(x^r_k, w^r_k) - A_kx^r_k - B_kw^r_k,
    \end{equation}
    where $A_k$ and $B_k$ are the system matrices, and $d_k$ is the residual term of the linearization.
    
    \subsection{Covariance-Controlled Trajectory Sampling}\label{Covariance controlled trajectory sampling}
    
    As with the baseline MPPI algorithm, the CC-MPPI algorithm simulates $M$ trajectories during each iteration. 
    Let $\U = \left[u_0,\ldots,u_{N-1}\right]^\intercal$ be the control sequence of the $m^{\text{th}}$ CC-MPPI sampled trajectory during the current iteration of the algorithm where we drop the superscript $m$ for simplicity. 
    The optimal control sequence from the previous iteration which is also the reference control sequence of the current iteration $\W = \left[w_{0}\,\ldots,w_{N-1}\right]^\intercal$ is injected with artificial noise 
    $\e = \left[\epsilon_0,\ldots,\epsilon_{N-1}\right]^\intercal$ and a feedback term $K_ky_k$ is added, such that
    \begin{equation}\label{eq11}
    u_k = w_k + \epsilon_k + K_ky_k, \qquad         \epsilon_k \sim \mathcal{N}(0,\Sigma_\epsilon),
    \end{equation}
    where $y_k$ follows the dynamics,
    \begin{equation}\label{eq12}
    y_{k+1} = A_ky_k + B_k\epsilon_k, \qquad y_0 = 0,
    \end{equation}
    where $y_0 = 0$ since we assume perfect observation of the initial state \cite{CSSMPC}. Substituting (\ref{eq11}) into (\ref{eq8}) yields,
    \begin{equation}\label{eq14}
    x_{k+1} = A_kx_{k} + B_k(w_{k} + K_ky_k) + d_k + B_k\epsilon_k,
    \end{equation}
    where $x_{k}$ is the state of the $m^{\text{th}}$ CC-MPPI sampled trajectory at step $k$, and $d_k$ is the residual term of the linearization at step $k$ as defined in (\ref{eq10}).
     Let $x_0$ be the state at the beginning of the current iteration.
     We can then rewrite the system in (\ref{eq14}) in the compact form,
    \begin{equation}\label{eq22}
    \X = \mathcal{A}x_0 + \mathcal{B}(\W+K \Y)+ \mathcal{C} \D+\mathcal{B} \e,
    \end{equation}
    where $\X = \left[x_0,\ldots,x_N\right]^\intercal \in \Re^{n_x (N+1)}$, $\W = \left[w_0, \ldots, w_{N-1} \right] \in \mathbb{R}^{n_u N}$
    $\Y = \left[y_0,\ldots,y_N\right]^\intercal \in \Re^{n_x (N+1)}$,
    $\D = \left[d_0,\ldots,d_{N-1}\right]^\intercal \in \Re^{n_x N}$, 
    $\e = \left[\epsilon_0,\ldots,\epsilon_{N-1}\right]^\intercal \in \Re^{n_x N}$
    and the augmented system matrices $\mathcal{A}$, $\mathcal{B}$, $\mathcal{C}$, $K$
    are defined similarly as in \cite{CSvehicle}. 
    In order to compute $K$ to satisfy the terminal covariance constraint (\ref{terminalconstraint}), the CC-MPPI considers a constrained LQG problem (\ref{eq26}) at each optimization iteration as follows
    \begin{subequations}\label{eq26}
    \begin{align}
        \begin{split}
       & \min_{K}J(\X,\W,K) = \\ &~~~\mathbb{E}\left[{\X}^\intercal\bar{Q}\X+(\W+K\Y)^\intercal\bar{R}(\W+K\Y)\right],\\
        \end{split}\\
    \text{subject to,} \nonumber\\
    &\X = \mathcal{A}x_0 + \mathcal{B}(\W+K\Y)+\mathcal{C} \D+\mathcal{B}\e,\\
    &E_N(\mathbb{E}\left[\X{\X}^\intercal\right] - \mathbb{E}\left[\X\right]\mathbb{E}\left[\X\right]^\intercal){E_N}^\intercal \preceq \Sigma_f, \label{eq26c}
    \end{align}
    \end{subequations}
    where $E_N = \left[ 0,\ldots,0,I \right]^\intercal \in \Re^{n_x \times n_x(N+1)}$, the augmented cost parameter matrices $\bar{Q}=\blkdiag(Q, \ldots, Q, Q_f)\in \mathbb{R}^{n_x (N+1) \times n_x (N+1)}$ and $\bar{R} = \blkdiag(R,\ldots,R) \in \mathbb{R}^{n_u N \times n_u N}$. Since  $\mathbb{E}\left[\e \right] = 0$ and $y_0 = 0$, we have $\mathbb{E}\left[\W+K\Y\right] = \W$. It follows from (\ref{eq12}) and (\ref{eq22}) that
    \begin{equation}\label{eq27}
    \bar{\X} = \mathbb{E}\left[ \X \right] = \mathcal{A}x_0 + \mathcal{B}\W + \mathcal{C}\D,
    \end{equation}
    and,
    \begin{equation}\label{eq28}
    \Tilde{\X} = \X - \mathbb{E}[\X]
    = \mathcal{B}K\Y + \mathcal{B}\e = (I+\mathcal{B}K)\mathcal{B}\e.
    \end{equation}
    The cost function $J(\X,\W,K)$ in (\ref{eq26}) can then be converted to the following equivalent form \cite{CS}
    \begin{equation}\label{eq29}
    \begin{split}
    J(\bar{\X}, \Tilde{\X}, &\W, K) = \textrm{tr}(\bar{Q}\mathbb{E} [ \Tilde{\X} {\Tilde{\X}}^\intercal ])\\
    & + {\bar{\X}}^\intercal \bar{Q} \bar{\X} + \textrm{tr}(\bar{R}\mathbb{E}\left[ K\Y \Y^\intercal K^\intercal \right]) + {\W}^\intercal \bar{R} \W.
    \end{split}
    \end{equation}
    \noindent The reference control sequence $\W$ is fixed and is given by the optimal control sequence from the previous CC-MPPI iteration, which implies that $\bar{\X}$ is fixed and is given by (\ref{eq27}). For the optimization problem in (\ref{eq26}), we can then drop the terms representing constant values in (\ref{eq29}) and obtain the cost
    \begin{equation}\label{eq31}
    J_{\Sigma}(\Tilde{\X}, K) = \textrm{tr}(\bar{Q}\mathbb{E} [ \Tilde{\X} {\Tilde{\X}}^\intercal ]) + \textrm{tr}(\bar{R}\mathbb{E}\left[ K\Y \Y^\intercal K^\intercal \right]).
    \end{equation}
     Substituting (\ref{eq28}) into (\ref{eq31}), and using the fact that $\mathbb{E}[\epsilon {\epsilon}^\intercal] = \Sigma_{\epsilon}$, yields,
    \begin{equation}\label{eq32}
    J_{\Sigma}(K) = \textrm{tr}\left(\left( (I+\mathcal{B}K)^\intercal\bar{Q}(I+\mathcal{B}K)+K^\intercal \bar{R}K \right)\mathcal{B}\bar{\Sigma}_{\epsilon}\mathcal{B}^\intercal \right),
    \end{equation}
    where $\bar{\Sigma}_{\epsilon} = \blkdiag(\Sigma_{\epsilon},...,\Sigma_{\epsilon}) \in \mathbb{R}^{n_u N \times n_u N}$. Substituting (\ref{eq22}) into (\ref{eq26c}), we obtain,
    \begin{equation}\label{eq34}
    \Sigma_f \succeq E_N(I+\mathcal{B}K)\mathcal{B}\bar{\Sigma}_{\epsilon}\mathcal{B}^\intercal(I+\mathcal{B}K)^\intercal E_N.
    \end{equation}
    Finally, Problem (\ref{eq26}) can be converted into the following convex optimization problem,
    \begin{equation}  \label{eq35}
    \min_K J_{\Sigma}(K) ~~\text{subject to}~~(\ref{eq34}).
    \end{equation}
    The problem (\ref{eq35}) can be easily solved by a convex optimization solver such as Mosek \cite{Mosek} to obtain $K$. Note that the last diagonal term $Q_f$ in $\bar{Q}$ can be viewed as a soft terminal state constraint that can be tuned to control the terminal covariance in situations where a hard constraint (\ref{eq34}) is not preferred.
    It follows from \eqref{eq11} that the control sequence of the $m^{\text{th}}$ sampled trajectory is $\U = \W + \e + K\Y$.
    We then rollout the sampled trajectories using $\U$ and the dynamical model (\ref{NonlinearSystem}). The complete CC-MPPI algorithm is detailed in Section \ref{The CC-MPPI Algorithm}.
    
\section{The CC-MPPI Algorithm}\label{The CC-MPPI Algorithm}
    
    The CC-MPPI algorithm is given in Algorithm \ref{algo1}. Line \ref{GetEstimateLine} obtains the current estimate of the state $x_{0}$ at the beginning of the current optimization iteration.  
    Lines \ref{RolloutBeginLine} to \ref{RolloutEndLine} rollout the reference trajectory $\X$ using the discrete-time nonlinear 
    dynamical model $F$. 
    Line~\ref{DynamicsLinearizationLine} linearizes the model $F$ along $\X$ and its corresponding control sequence $\W$ as described in (\ref{eq7}), (\ref{eq8}), (\ref{eq9}), (\ref{eq10}), and calculates the augmented dynamical model matrices $\mathcal{A}$, $\mathcal{B}$, $\mathcal{C}$ along with the linearization residual term $\D$.
    Line \ref{CovarianceControllLine} computes the feedback gain $K$ for the closed-loop system in (\ref{eq22}) by solving the convex optimization problem (\ref{eq35}). 
    Lines \ref{TrajectorySamplingBeginLine} to \ref{TrajectorySamplingEndLine} sample the control sequences, perform the rollouts and evaluate the sampled trajectories with the running cost \eqref{ModifiedCost}. 
    Specifically, lines \ref{ControlandUncontrolledCommandsBeginLine} to \ref{ControlandUncontrolledCommandsEndLine} introduce sample trajectories of the close-loop dynamics and sample trajectories of zero-mean input, so that the algorithm can balance between smoothness of trajectories and low control cost  \cite{MPPI_Theory_and_Applications_to_Autonomous_Driving}. 
    Line \ref{CalculateOptimalControlLine} computes the optimal control sequence $\W$ following \eqref{eq3} and \eqref{eq5}. 
    Line \ref{ExecuteCommandLine} sends the first control command $w_{0}$ of the optimal control sequence to the actuators. 
    Line \ref{InitializationLine} removes $w_{0}$, and duplicates $w_{N-1}$ at the end of the horizon for $\W$.\\
    
    \begin{algorithm}[h]
        \caption{CC-MPPI Algorithm}\label{algo1}
        \SetAlgoLined
        \LinesNumbered
        \SetKwInOut{Input}{Given}
        \Input{
        \noindent $\gamma, \lambda, \Sigma_{\epsilon}, \Sigma_{f}, \alpha: \text{CC-MPPI Parameters}$;}
        \SetKwInOut{Input}{Input}
        \Input{
        \noindent $\W :\text{Initial control sequence}$}
        
        \While{task not complete}{
            $x_{0} \leftarrow \textit{GetStateEstimate}()$;\\ \label{GetEstimateLine}
            \For{$k \leftarrow 0 \textbf{ to } N-1$\label{RolloutBeginLine}}
            {
            $x_{k+1} \leftarrow F(x_{k}, w_{k})$;\\
            }\label{RolloutEndLine}
            $\mathcal{A}, \mathcal{B}, \mathcal{C}, \D \leftarrow \textit{DynamicsLinearization}(F,\X,\W);$\\ \label{DynamicsLinearizationLine}
            $K = \textit{CovarianceControl}(\Sigma_f,\Sigma_\epsilon, \mathcal{B}, \bar{Q}, \bar{R})$; \\ \label{CovarianceControllLine}
            \For{$m\leftarrow 1 \textbf{ to } M$}{\label{TrajectorySamplingBeginLine} 
                  $x^{(m)}_{0}\leftarrow x_0,\quad y^{(m)}_{0}\leftarrow 0;$\\
                  $\text{Sample }\e^{(m)} \leftarrow \left[\epsilon^{(m)}_0,\ldots,\epsilon^{(m)}_{N-1}\right]^\intercal$;\\
                  \For{$k\leftarrow 0 \textbf{ to } N-1$}{
                  \uIf{$m < (1-\alpha)M$}{ \label{ControlandUncontrolledCommandsBeginLine}
                  $v^{(m)}_{k} \leftarrow w_{k} + K_k y^{(m)}_k$;
                  }\Else{
                  $v^{(m)}_{k} \leftarrow 0$\\
                  }
                  $u^{(m)}_{k} \leftarrow v^{(m)}_{k} + \epsilon^{(m)}_k$;\\
                  $x^{(m)}_{k+1} \leftarrow F(x^{(m)}_k , u^{(m)}_k)$;\\\label{ControlandUncontrolledCommandsEndLine}
                  $y^{(m)}_{k+1} \leftarrow A_k y^{(m)}_k + B_k \epsilon^{(m)}_k$;\\
                  $S_{m} \leftarrow S_{m} + \Tilde{q}(x^{(m)}_k, v^{(m)}_k, \epsilon^{(m)}_k)$;\\
                  }
                  $S_m \leftarrow S_{m} + \phi(x^{(m)}_N)$;\\
            } \label{TrajectorySamplingEndLine}
            $\W = \textit{CalculateOptimalControl}(S_m, \U^{(m)})$;\\\label{CalculateOptimalControlLine}
            $\textit{ExecuteCommand}(w_{0})$;\\ \label{ExecuteCommandLine} 
            $\textit{Initialize}(\W)$;\label{InitializationLine}
        }
    \end{algorithm}
\section{Results}\label{Numerical Simulations Section}

    In this section, we show via a series of numerical examples that the proposed CC-MPPI algorithm outperforms the baseline MPPI algorithm in critical situations described in Section~\ref{Problem Formulation}. 
    The terminal covariance $\Sigma_f$ in (\ref{eq34}) for the CC-MPPI should be determined based on the environment. 
    Please also refer to the accompanied video\footnote{\url{https://www.youtube.com/watch?v=cZq4tnBTIqc}} for a demonstration of more simulation examples.

\subsection{Vehicle Model}

    We assume that the injected artificial noise $\epsilon^{(m)}$ in CC-MPPI and MPPI algorithms is significantly greater than the inherent noise of the vehicle model, such that the model noise is negligible. 
    We model the vehicle using a single-track bicycle model
    \begin{subequations}  \label{bicycle}
    \begin{align}
    \dot{x} = v \cos(\beta + \phi),&\quad
    \dot{y} = v \sin(\beta + \phi),\\
    \dot{\phi} = \frac{v}{l_r \sin \beta },&\quad
    \dot{v} = \delta_T ,
    \end{align}
    \end{subequations}
    where $\tan \beta = {l_r}/{(l_f+l_r)} \tan \delta_s$ 
    and the parameters $l_r$, $l_f$ are distances from the CoM to the rear and front wheels, respectively. 
    The $x$, $y$ are position coordinates of a fixed world coordinate frame. 
    The $\phi$ is the vehicle yaw angle, and $v$ is velocity at CoM with respect to the world coordinate frame. The $\delta_T$ and $\delta_s$ are throttle and steering inputs to the model, respectively. 
    We discretize the system \eqref{bicycle} with the Euler method, $x_{k+1} = F(x_k, u_k) = x_k + f(x_k, u_k) \Delta t$, where $k = 0,\ldots,N-1$ and time step $\Delta t = 0.02~\textrm{s}$.
    
\subsection{Controller Setup}
    
\begin{figure}[t]
    \centerline{\includegraphics[scale = 0.25]{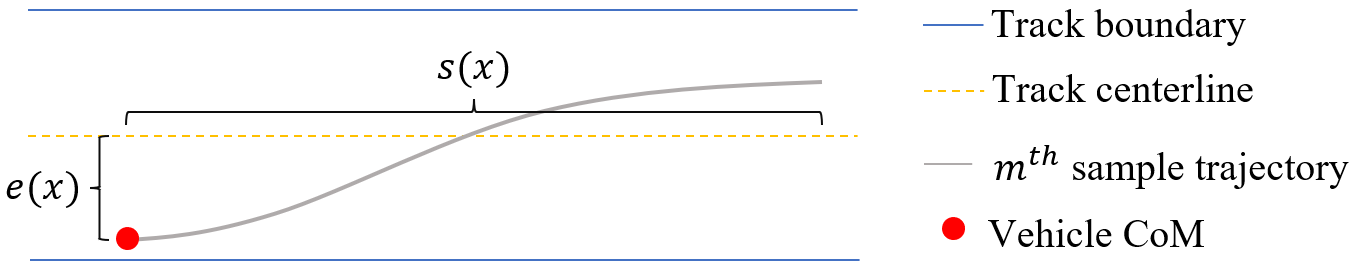}}
    \caption{Track schematic}
    \label{TrackSchematics}
    \end{figure}

    Assuming that the model noise is much smaller than the injected noise $\epsilon^{(m)}$,
    it follows that $\nu^{-1}\approx 0$ in (\ref{ModifiedCost}). 
    It then follows from (\ref{ModifiedCost}) that the running cost $\Tilde{q}(x^{(m)}_k, v^{(m)}_k, \epsilon^{(m)}_k)$ 
    for the $k^{\textrm{th}}$ step of the $m^{\textrm{th}}$ sampled trajectory takes the form,
    \begin{equation}\label{RunningCost}
    \begin{split}
    &\Tilde{q}(x^{(m)}_k, v^{(m)}_k, \epsilon^{(m)}_k) = q(x^{(m)}_k) + \\
    &\frac{1}{2}{\epsilon^{(m)}_k}^\intercal R \epsilon^{(m)}_k + {v^{(m)}_k}^\intercal R \epsilon^{(m)}_k + \frac{1}{2}{v^{(m)}_k}^\intercal R {v^{(m)}_k},
    \end{split}
    \end{equation}
    for $k = 0,\ldots,N-1$,  $m = 1,\ldots,M,$
    where we take the state-dependent cost $q(x_k^{(m)})$ as,
    \begin{equation}\label{state_dependent_cost}
        q(x)=\mu_{\textrm{bdry}}(x) + c_1  \mu_{\textrm{obs}}(x).
    \end{equation}
    The term $\mu_{\textrm{bdry}}(x)$ in \eqref{state_dependent_cost} is the boundary cost which prevents the vehicle from leaving the track, and it is given by
    \begin{equation}\label{BoundaryCost}
        \mu_{\textrm{bdry}}(x)=\left\{
                \begin{array}{ll}
                  0, \quad \textrm{if } \textit{WithinBoundary}(x),\\
                  2000, \quad \textrm{Otherwise}.
                \end{array}
              \right.
    \end{equation}
    The term $c_1 \mu_{\textrm{obs}}(x)$ in \eqref{state_dependent_cost} penalizes collisions with obstacles, where $c_1$ is a weighting coefficient.
    We choose two different forms of $\mu_{\textrm{obs}}(x)$ in our simulations. 
    The first is discontinuous on the obstacles' edges, 
    \begin{equation}\label{ObstacleCostDiscontinuous}
        \mu_{\textrm{obs}}(x)=\left\{
        \begin{array}{ll}
          0, \quad \textrm{if } r_i > d_i(x),\\
          10, \quad \textrm{Otherwise},
        \end{array}
      \right.
    \end{equation}
    and the second is continuous on the obstacles' edges,
    \begin{equation}\label{ObstacleCostContinuous}
        \mu_{\rm obs}(x)= \sum^P_{i=1} \max(r_i - d_i(x), 0),
    \end{equation}
    where $d_i(x)$ describes the distance from the vehicle's CoM to the center of the $i\textrm{th}$ circular obstacle, and $r_i$ is the radius of the $i\textrm{th}$ obstacle. 
    We take $r_i = 0.1$ m for all of the $P$ circular obstacles in this section. 
    In our simulations, the terminal cost $\phi(x^{(m)}_N)$ for the MPPI and CC-MPPI controllers has the form,
    \begin{equation}\label{terminal_cost}
        \phi(x) = c_2  (1-s(x)) + 500  e^2(x).
    \end{equation}
    The first term in \eqref{terminal_cost} is the progress cost, where $c_2$ is a weighting coefficient and $s(x)$ represents the distance between the current vehicle state and the terminal state of the sample trajectory along the track centerline. 
    The term $e(x)$ in \eqref{terminal_cost} is the vehicle's lateral deviation from the track centerline (see Figure~\ref{TrackSchematics}).
    For both the MPPI and the CC-MPPI controllers, we set the control horizon to $N = 15$, the value for $\lambda = 1$, the number of sampled trajectories to $M = 4096$, the portion of uncontrolled sampled trajectories to $\alpha = 0.2$, and the control cost matrix to $R = 0.01 I$.
    The parameter values discussed here are shared by all the controller setups in the simulations of this section.\\
    
    \subsection{Planning in Fast-changing Environment: Unpredictable Obstacles}
    
\begin{figure}[t]
    \centerline{\includegraphics[scale = 0.28]{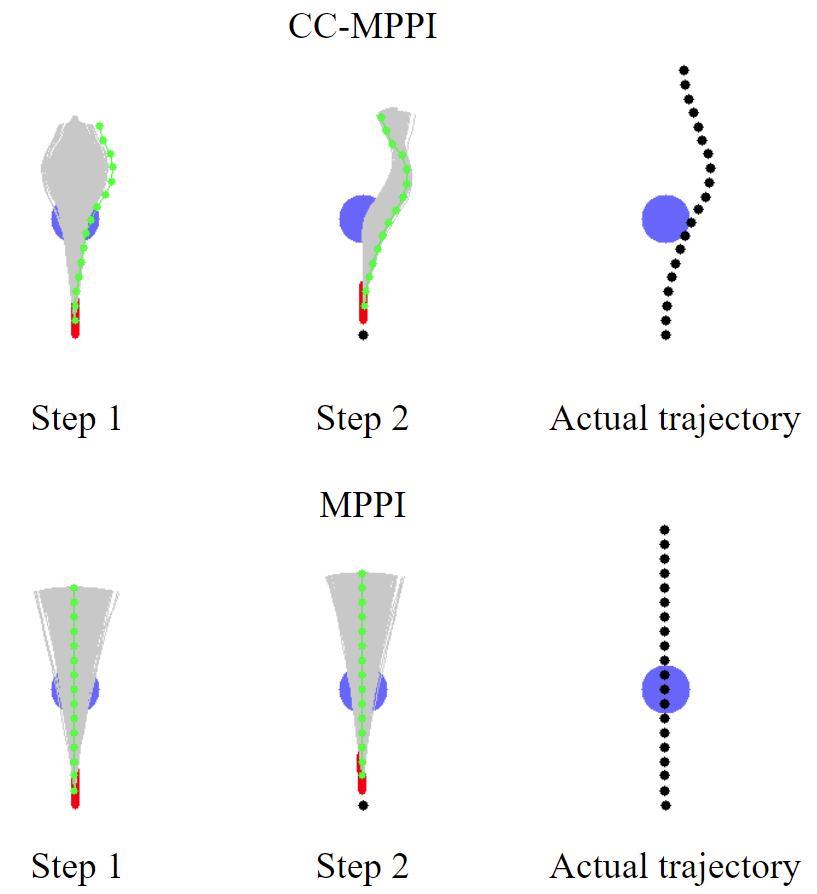}}
    \caption{Responses of the MPPI and CC-MPPI to an unpredictable obstacle.The grey curves are the sampled trajectories, the green curves represent the predicted optimal trajectories generated by the controllers, and the black points show the actual trajectories taken by the vehicle.}
    \label{Responses of the MPPI and CC-MPPI to Emergency}
    \end{figure}

    This experiment tests the CC-MPPI controller's ability to respond to emergencies owing to unpredictable appearance of obstacles. 
    We test the CC-MPPI against a baseline MPPI controller in an environment where an obstacle suddenly appears in the traveling direction of an autonomous vehicle. 
    In this simulation, the CC-MPPI and MPPI controllers were injected with noises having the same covariance $\Sigma_{\epsilon} = \diag(0.0049, 0.0012)$, and the same weighting coefficients $c_1 =712.5$ and $c_2=3.3$ in their trajectory costs.
    Figure~\ref{Responses of the MPPI and CC-MPPI to Emergency} demonstrates that the MPPI controller fails to find a feasible solution and results in a collision with the obstacle. 
    Figure~\ref{Responses of the MPPI and CC-MPPI to Emergency} further shows that the CC-MPPI 
   has a more effective trajectory sampling distribution strategy, 
   which leads the vehicle to take a feasible trajectory that avoids collision.
    
    \subsection{Aggressive Driving in a Cluttered Environment}
    
    To further examine the performance of the CC-MPPI controller in more complicated environments, we run simulations using a CC-MPPI controller and an MPPI controller on a race track environment with obstacles densely scattered on the track. The track has a constant width of 0.6~m, the centerline has a length of 10.9~m and each turn of the centerline has radius 0.3 m. 
    Each obstacle has radius 0.1~m and uses the continuous obstacle cost \eqref{ObstacleCostContinuous}. 
    Both controllers are set to achieve minimum lap time while avoiding collisions with cluttered obstacles, and have the same covariance $\Sigma_{\epsilon} = \diag(0.49, 0.12)$ for their injected noises. 
    We perform a grid search by varying the cost weight $c_1$ in \eqref{state_dependent_cost} which corresponds to avoiding collisions with obstacles, and the cost weight $c_2$ in \eqref{terminal_cost} for optimizing the vehicle velocity along the track centerline. 
    Table~\ref{grid_search} shows the grid search parameters. 
    Figure~\ref{pareto} presents the results of the grid search in a scatter plot showing the distribution of lap times and average number of collisions. Table \ref{MPPI and CCMPPI statistics} summarizes the grid search and shows the statistics over all the simulations. 
    
    \begin{table}[t]
    \caption{CC-MPPI and MPPI Grid Search Parameter Values}
    \begin{center}
    \begin{tabular}{|c|c|c|c|}
    \hline
    \textbf{Cost Parameter} & \textbf{{Min}}& \textbf{{Max}} & \textbf{{Interval}} \\
    \hline
    $c_1$      & 75 & 450 & 37.5 \\
    \hline
    $c_2$      & 1.65 & 2.97 & 0.33   \\
    \hline
    \end{tabular}
    \label{grid_search}
    \end{center}
    \end{table}
    
    \begin{figure}[t]
    \centerline{\includegraphics[scale = 0.8]{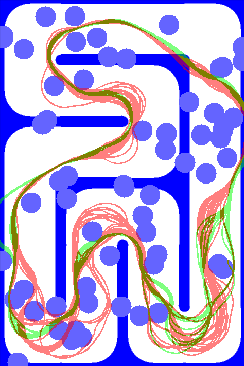}}
    \caption{MPPI and CC-MPPI trajectories on a race track. 
    The trajectories in red are generated by the CC-MPPI, and the trajectories in green are generated by the MPPI using the same injected noise covariance as the CC-MPPI.}
    \label{track_MPPI_CCMPPI}
    \end{figure}
    
    We define a failure to be the situation when the vehicle comes to a complete stop, or when the vehicle is too far away from the track centerline ($ >1$~m). 
    If the vehicle finishes $20$ laps without a failure, the simulation is considered  a success.
    Figure~\ref{pareto} shows that the data points corresponding to CC-MPPI occupy the bottom part of the scatter plot, which indicates that the CC-MPPI generates trajectories that are significantly faster than those by the MPPI controller. 
    Table~\ref{MPPI and CCMPPI statistics} shows that the CC-MPPI achieves $34.78\%$ smaller average lap time, $6.34\%$ fewer collisions and $218.97 \%$ higher success rate than the MPPI in simulations. 
    Moreover, the two data points in the red circles in Figure~\ref{pareto} are produced by MPPI and CC-MPPI with the same set of $c_1$ and $c_2$ values, and Figure~\ref{track_MPPI_CCMPPI} visualizes the trajectories that correspond to these two data points. 
    We see that the CC-MPPI generates a more aggressive driving maneuver than the MPPI, which helps explain why the CC-MPPI achieves a significantly smaller average lap time.
    
    The performance of CC-MPPI, however, comes with an increased computational overhead. 
    Using our implementation, the CC-MPPI controller runs at 13Hz, while the MPPI controller runs at 97Hz. 
    All simulations were done on a desktop computer equipped with an i9 3.5GHz CPU, and an RTX3090 GPU. 
    The main computational bottleneck of CC-MPPI is the computation of the feedback gain $K$ at each iteration.
    Possible remedies include updating the feedback gain asynchronously, computing the feedback gains off-line and storing them in a lookup table,
    or using a faster, dedicated 
    convex optimization solver that is more suitable for real-time implementation~\cite{Mao2018,mattingley2012cvxgen,dueri2014automated}.
    
    \begin{table}[t]
    \caption{CC-MPPI Vs. MPPI of different settings}
    \label{MPPI and CCMPPI statistics}
    \begin{center}
    \begin{tabular}{ | c | c | c | c |}
    \hline
    \textbf{Controller} & \textbf{{Avg. laptime(s)}} & \textbf{{No. collision/lap}} & \textbf{{Success rate}} \\ \hline
    {CC-MPPI} & 4.20 & 2.68 & 98.18\% \\ \hline
   {MPPI} & 6.44 & 2.86 & 30.78\% \\ \hline
    \end{tabular}
    \end{center}
    \end{table}
    
    \begin{figure}[t]
    \centerline{\includegraphics[scale = 0.45]{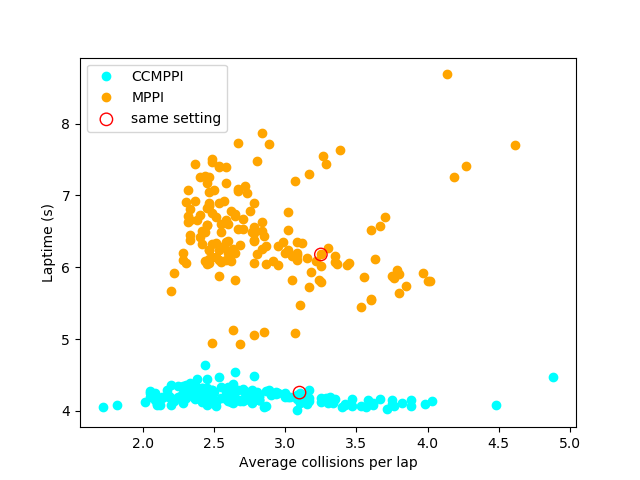}}
    \caption{Lap time and number of collisions distribution. 
    Each point in this figure shows the average lap time and average number of collisions over 20 laps in a simulation using one pair of $c_1$ and $c_2$ from Table \ref{grid_search}. The orange points are produced by the MPPI controller and the cyan points are by the CC-MPPI controller.}
    \label{pareto}
    \end{figure}
    
\section{Conclusions And Future Work}

    We have proposed the Covariance-Controlled Model Predictive Path Integral (CC-MPPI) algorithm that incorperates covariance steering within the MPPI algorithm. 
    The CC-MPPI algorithm has adjustable trajectory sampling distributions which can be tuned by changing the terminal covariance constraint $\Sigma_f$ in (\ref{terminalconstraint}) and the covariance of the injected noise $\Sigma_{\epsilon}$ in (\ref{eq1c}), which makes it more flexible and robust than the MPPI algorithm.
    In the simulations, we showed that the CC-MPPI explores the environment and samples trajectories more efficiently than MPPI for the
    same level of exploration noise ($\Sigma_{\epsilon}$).
    This results in the vehicle responding faster to unpredictable obstacles and avoid collisions better in a cluttered environment than MPPI. 
    The CC-MPPI performance can be further improved if $\Sigma_{\epsilon}$ and $\Sigma_f$ are tuned based on the information of the surrounding environment.
    
    In the future, we plan to design policies to choose judiciously the terminal covariance constraint $\Sigma_f$ and  the injected noise covariance $\Sigma_{\epsilon}$ on-the-fly. 
    These policies should evaluate the environment and assign $\Sigma_f$, $\Sigma_{\epsilon}$ for the CC-MPPI controller, such that the trajectory sampling distribution of the controller can be tailored to carry out informed and efficient sampling in any environments.

\bibliographystyle{IEEEtran}
\bibliography{bib/references}

\end{document}